\pdfoutput=1

\documentclass[11pt]{article}
\usepackage[final]{ACL2023}
\usepackage{times}
\usepackage{latexsym}
\usepackage[T1]{fontenc}
\usepackage[utf8]{inputenc}
\usepackage{microtype}
\usepackage{inconsolata}
\usepackage{booktabs}
\usepackage{siunitx}
\usepackage{times}
\usepackage{tipa}
\usepackage{latexsym}
\usepackage{multirow}
\usepackage[table]{xcolor}
\usepackage{subcaption}
\usepackage{amsmath}
\usepackage{amsfonts}
\usepackage{amssymb}
\usepackage[symbol]{footmisc}
\usepackage{todonotes}
\usepackage{l  ongtable}
\usepackage{array} 
\usepackage{ragged2e} 
\newcommand{\nexttablenum}{\number\numexpr\value{table}\relax}

\title{\textit{ASR Under Noise}:
  Exploring Robustness for Sundanese and Javanese}

\author{
  \textbf{Salsabila Zahirah Pranida}$^{*,1}$ 
  \textbf{Muhammad Cendekia Airlangga}$^{*,1}$ 
  \textbf{Rifo Ahmad Genadi}$^{*,1}$ \\
  \textbf{Shady Shehata}$^{2}$ \\
  $^{1}$ MBZUAI \quad $^{2}$ University of Waterloo \\
  \texttt{\small \{salsabila.pranida, muhammad.airlangga, rifo.genadi\}@mbzuai.ac.ae}
  \\
  \small $^{*}$ Equal contribution 
}

\begin{document}
\maketitle
\begin{abstract}
We investigate the robustness of Whisper-based automatic speech recognition (ASR) models for two major Indonesian regional languages: Javanese and Sundanese. While recent work has demonstrated strong ASR performance under clean conditions, their effectiveness in noisy environments remains unclear. To address this, we experiment with multiple training strategies, including synthetic noise augmentation and SpecAugment, and evaluate performance across a range of signal-to-noise ratios (SNRs). Our results show that noise-aware training substantially improves robustness, particularly for larger Whisper models. A detailed error analysis further reveals language-specific challenges, highlighting avenues for future improvements. Code is available at \url{https://github.com/rifoagenadi/robust_jvsu_asr}.
\end{abstract}

\section{Introduction}

Automatic Speech Recognition (ASR) systems have made remarkable progress in recent years, especially for high-resource languages like English. While modern ASR handles diverse accents~\citep{7953071} and noise~\citep{6639100} in high-resource languages, it remains unreliable for low-resource ones.

Indonesia, with 284M people and over 700 languages, is among the world’s most linguistically diverse countries~\citep{bps2025statistik,ethnologue2025, kemdikbudPetaBahasa, bpssukubudaya2024}. Yet, both remain underrepresented in ASR research and resources.

\begin{figure*}[ht]
    \centering    \includegraphics[width=\linewidth]{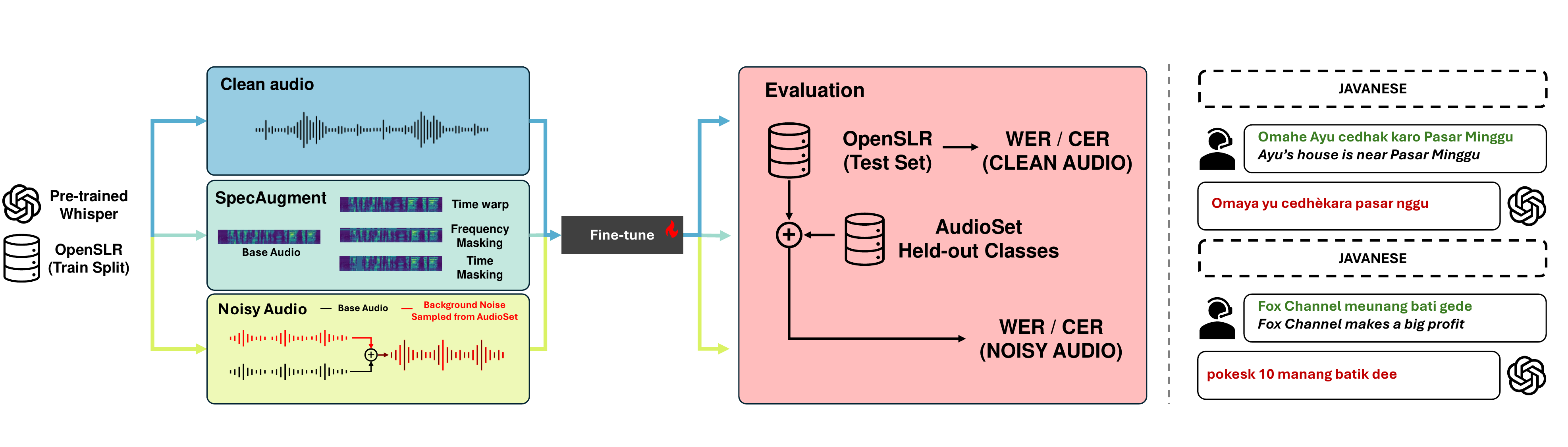}
    \caption{\textbf{(Left)} Training and evaluation pipeline for Whisper-based ASR models. Each fine-tuned model is evaluated on clean and noisy versions of the OpenSLR test set.\textbf{ (Right)} Examples of noisy transcriptions in Javanese and Sundanese using Whisper. 
    The top boxes show spoken utterances with noise; the bottom boxes show the corresponding ASR outputs, demonstrating significantly degraded quality under noisy conditions.}
    \label{fig:experimental-pipeline}
\end{figure*}

These languages exhibit high dialectal variation and are spoken daily in uncontrolled, noisy settings, which makes them difficult for standard ASR models, which are mostly trained on Indo-European data~\citep{towards2012vowel}. Figure~\ref{fig:experimental-pipeline} right illustrates how background noise severely degrades transcription quality, even with advanced models like Whisper. This demonstrates the vulnerability of current ASR systems to real-world acoustic challenges.

Amid the growing use of large-scale speech-language models, Whisper has emerged as a strong multilingual ASR system~\citep{radford2023robust}. Unlike prior models such as wav2vec 2.0 and XLS-R, Whisper demonstrates superior robustness and generalization, particularly in noisy and low-resource scenarios~\citep{pratama2024whisper, shah2024speechrobustbenchrobustness}. These strengths make Whisper an ideal foundation for exploring ASR robustness in Javanese and Sundanese.

In this work, we present the first systematic study of ASR robustness to noise in these languages using over 60 hours of training data. Our key takeaways are: (1) evaluating Whisper models across clean and noisy test conditions; (2) exploring training strategies like \texttt{SpecAugment} and noise-aware fine-tuning; (3) analyzing language-specific transcription errors; and (4) releasing our training and evaluation pipeline for reproducibility. This is the first work to benchmark ASR robustness to noise in these languages systematically.

\section{Related Works}

\paragraph{ASR for Sundanese and Javanese}

The \texttt{NusaASR} benchmark~\cite{cahyawijaya-etal-2023-nusacrowd} evaluates ASR models on Javanese and Sundanese primarily in zero-shot settings. While prior work has fine-tuned large models like XLS-R and Whisper~\cite{arisaputra2024xls, pratama2024whisper}, these efforts often rely on limited data and lack reproducibility. Moreover, they rarely address robustness under noisy conditions. In contrast, our work provides a more comprehensive evaluation by fine-tuning Whisper across both languages.

\paragraph{Noise Robustness}
Ensuring ASR robustness in noisy environments is a well-recognized challenge~\cite{shah2024speechrobustbenchrobustness, Feng2021ASRGLUEAN, Likhomanenko2020RethinkingEI}. Prior work addresses this through data augmentation techniques such as synthetic noise injection and room impulse responses. Among these, \texttt{SpecAugment}~\cite{park2019specaugment} has gained popularity as a simple and effective method. Other approaches include noise-aware training~\cite{orel2023noise} and denoising front-ends~\cite{dissen2024enhanced}. In our work, we independently evaluate \texttt{SpecAugment} and noise-aware fine-tuning, using noise samples from \texttt{AudioSet}~\cite{audioset}, as two distinct strategies to improve ASR robustness.

\begin{figure*}[ht]
    \centering
    \begin{subfigure}[t]{0.48\textwidth}
        \centering
        \includegraphics[width=\linewidth]{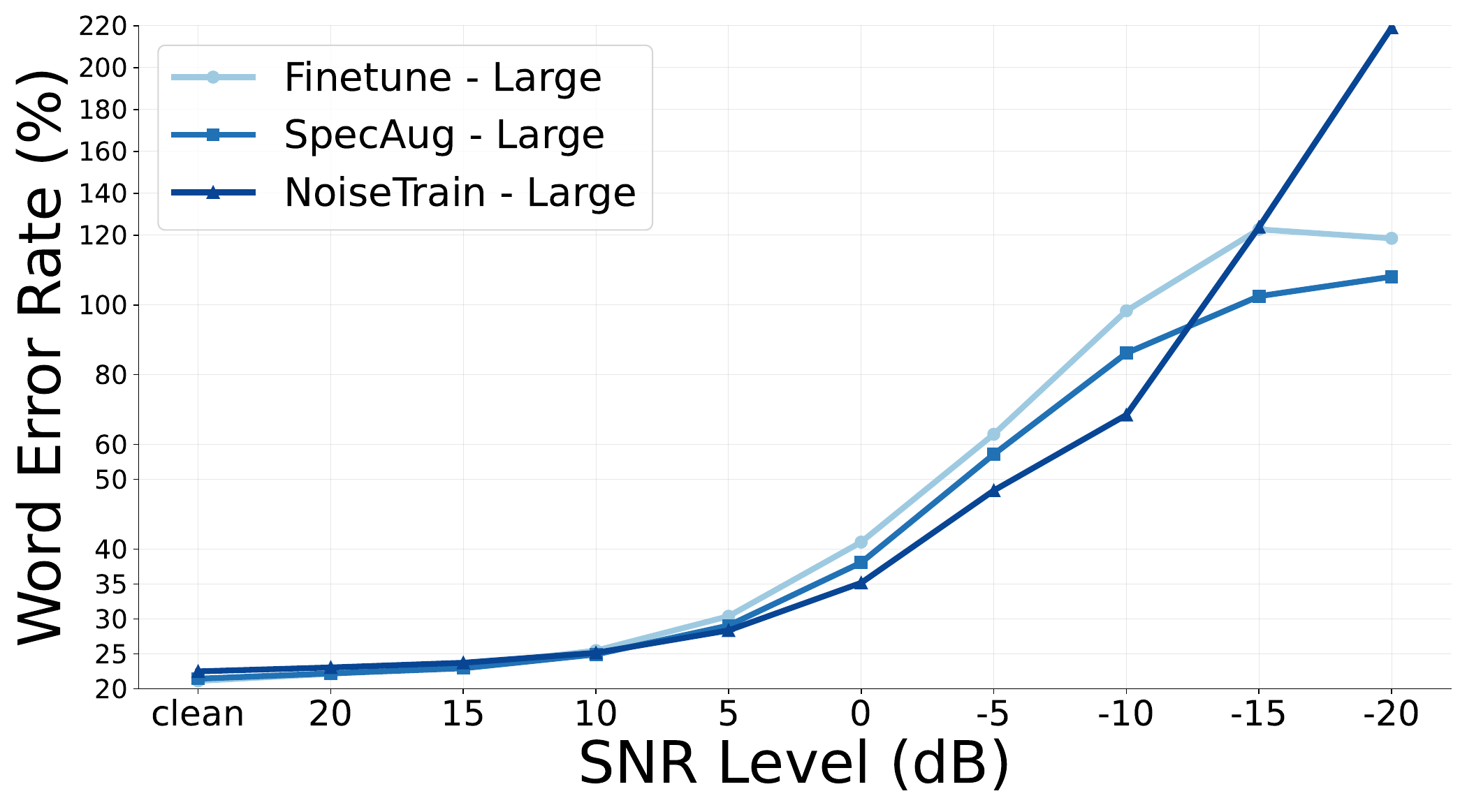}
        \caption{Javanese}
        \label{fig:choosen-large-jv}
    \end{subfigure}
    \hfill
    \begin{subfigure}[t]{0.48\textwidth}
        \centering
        \includegraphics[width=\linewidth]{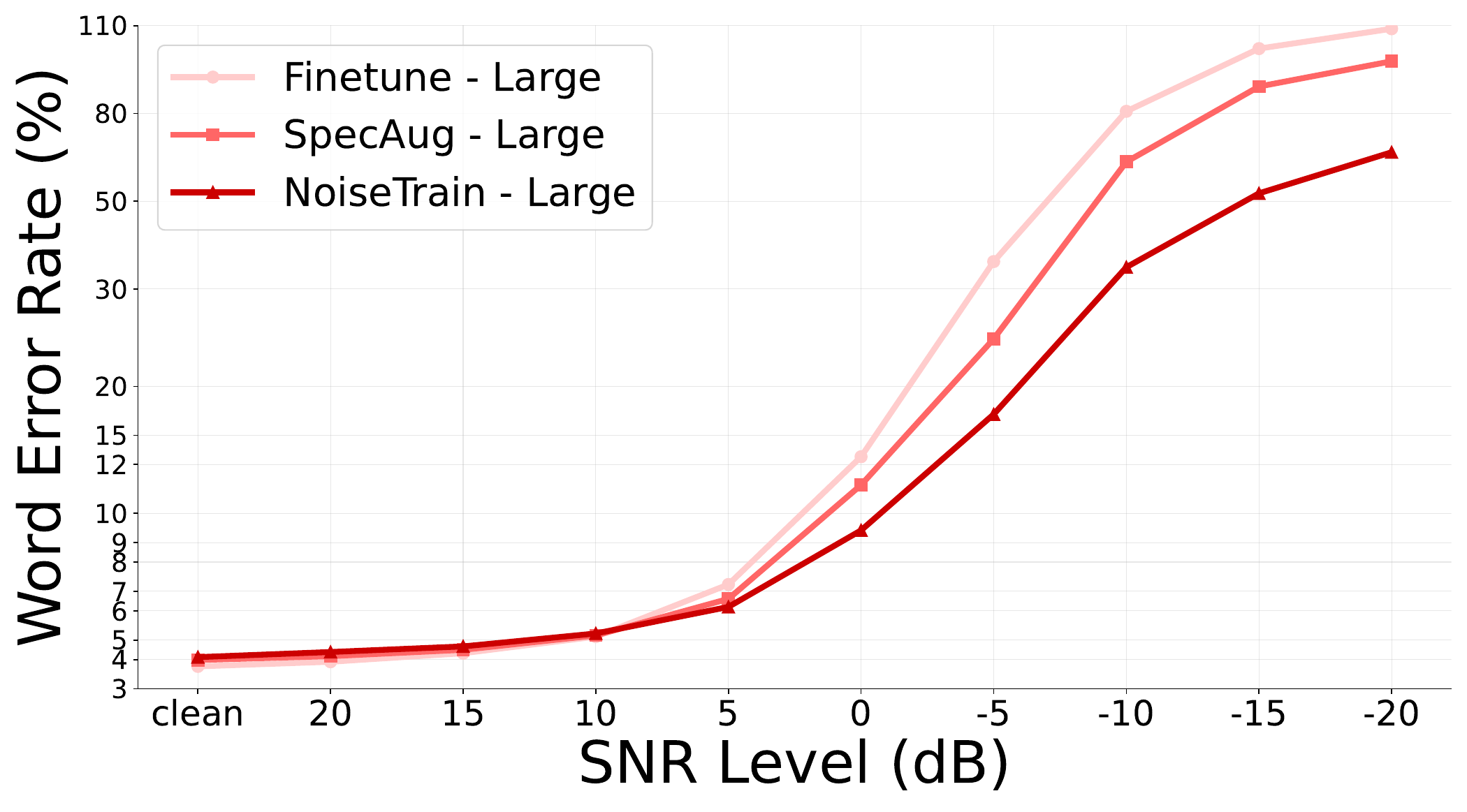}
        \caption{Sundanese}
        \label{fig:choosen-large-su}
    \end{subfigure}
    \caption{WER performance of Large-v3 Whisper across different SNR levels for Javanese and Sundanese. Models trained with \texttt{NoiseTrain} consistently outperform others under low-SNR conditions. Higher SNR values indicate cleaner audio.}
    \label{fig:whisper-wer-grid}
\end{figure*}

\section{Experimental Setup}
\label{sec:experimental-setup}

\subsection{Linguistic Characteristics}
\paragraph{Javanese}
Javanese has more than 80 million speakers~\citep{eberhardEthnologue2021} and is part of the Austronesian, Malayo Polynesian family~\citep{cohn2014local}. It is agglutinative with extensive affixation that produces many word forms and is commonly divided into Western, Central, and Eastern varieties, each with distinct phonology and vocabulary~\citep{wedhawati2001tata}. A notable feature is its speech levels, such as \textit{ngoko} (informal) and \textit{krama} (polite), which encode social hierarchy in interaction~\citep{isodarus2020penggunaan}. 

\paragraph{Sundanese}
Sundanese, spoken by about 30–40 million people in western Java~\citep{eberhardEthnologue2021}, is part of the Austronesian, Malayo Polynesian family and shows agglutinative morphology with rich affixation. Major dialects include Bogor, Priangan, and Cirebon, which differ in vocabulary and pronunciation~\citep{kurniawan2013sundanese}. The language also encodes politeness through registers that guide lexical choice.

\subsection{Dataset}
\paragraph{Data Overview} 
We use the OpenSLR Javanese and Sundanese corpora~\citep{kjartansson2018crowd}, collected with support from Universitas Gadjah Mada in Yogyakarta and Universitas Pendidikan Indonesia in Bandung. The recordings are read speech from volunteers. These corpora are valuable but do not cover the full range of dialects or spontaneous use.

From the full releases (185k utterances / 296 hours for Javanese and 219k utterances / 333 hours for Sundanese), we selected 10 subsets for training and 6 for testing~\citep{kjartansson2018crowd}. This gives about 60 hours of training data and 10 hours of test data per language, with train and test speakers kept separate (Table~\ref{tab:utterance_stats}). The size is adequate for baseline ASR, but limited coverage should be considered when interpreting results. While we were unable to identify detailed dialectical or speaker variations from the original paper \citet{kjartansson2018crowd}, we estimated the proportion of female and male speakers using a fine-tuned version of wav2vec \citep{baevski2020wav2vec20frameworkselfsupervised}\footnote{https://huggingface.co/prithivMLmods/Common-Voice-Gender-Detection}.

\begin{table}[ht]
\centering
\begin{tabular}{lccc}
\toprule
\textbf{Lang} & \textbf{Train} & \textbf{Test} & \textbf{\#Speakers (F\%)} \\
\midrule
JV  & 37,439 & 6,276 & 758 (57\%) \\
SU & 39,560 & 6,563 & 529 (57\%) \\
\bottomrule
\end{tabular}
\caption{Number of utterances and unique speakers for each language, with female speaker proportion.}
\label{tab:utterance_stats}
\end{table}

\paragraph{Synthetic Noise Data Generation} 
To simulate real-world conditions, we augment clean training data with background noise at various Signal-to-Noise Ratio (SNR) levels, following prior work~\cite{orel2023noise,maas2012recurrent}. The noise types reflect common environments like traffic and indoor chatter. Details on the noise selection, SNR values, and mixing procedure are provided in Appendix~\ref{appendix:noise-generation}.

\subsection{Training Pipeline}

We fine-tune four Whisper variants—\texttt{Tiny}, \texttt{Medium}, \texttt{Large-v3}, and \texttt{Large-v3-Turbo}—on Javanese and Sundanese ASR using OpenSLR. While these models support the languages, their zero-shot performance is poor due to limited training exposure. We explore three training strategies to improve robustness, as illustrated in Figure~\ref{fig:experimental-pipeline}.

\paragraph{Clean Fine-tuning} Models are trained on unmodified OpenSLR data as a baseline.

\paragraph{Clean + SpecAugment}
In this setup, we fine-tune the models by applying \texttt{SpecAugment} on clean data, a data augmentation method that applies time and frequency masking on input spectrograms. To tune augmentation hyperparameters, we use a 90/10 split of the training data for training and validation (see details in Appendix~\ref{appendix:experiments}). 

\paragraph{Fine-tune in Noisy Audio}
We synthetically augment the training set by mixing clean OpenSLR utterances with background sounds from 24 classes in \texttt{AudioSet}~\cite{audioset}, at various SNR levels. Noise audio in the train splits is shuffled and mapped in a many-to-one manner to SNR values. It means that one SNR was used for different audio files, but the audio files did not repeat. The resulting noisy dataset is then used to fine-tune the Whisper models. This setup is referred to as \texttt{NoiseTrain}.

\subsection{Evaluation Pipeline}

Models are evaluated on both clean and synthetic noisy versions of the OpenSLR test set, as shown on the evaluation side of Figure~\ref{fig:experimental-pipeline}, using word error rate (WER) as the main metric. Noisy test sets are created by mixing the clean utterances with background sounds from 8 held-out noise \texttt{AudioSet} classes\footnote{See Appendix~\ref{appendix:audioset} for the list of held-out noise classes.}.

\section{Results and Analysis}
\subsection{Model Robustness}

We evaluate Whisper models on Javanese and Sundanese under varying noise conditions. Figure~\ref{fig:whisper-wer-grid} shows how WER changes across SNR conditions using the \texttt{Large-v3} model (see details in Appendix~\ref{appendix:snr-results}), while Tables~\ref{tab:wer-javanese} and~\ref{tab:wer-sundanese} report detailed results for all model variants and training strategies. Zero-shot performance is poor, with WERs exceeding 70–120 even on clean audio, confirming that adaptation is critical. We selected \texttt{SpecAugment}  configuration \textbf{\#9} as the best-performing setup (see Appendix~\ref{appendix:experiments}) and use it for all reported results. Both \texttt{NoiseTrain} and \texttt{SpecAugment} significantly improve robustness, especially under low-SNR conditions. 

Models trained with \texttt{NoiseTrain} or \texttt{SpecAugment} consistently outperform clean-only models, especially under low-SNR conditions. For instance, in Javanese --SNR, \texttt{Medium} improves from 225.38 to 111.89 WER, and in Sundanese, from 199.09 to 56.15. Even larger models like \texttt{Large-v3} benefit, dropping from 79.91 to 41.37, showing the importance of noise-aware training for real-world robustness. Running all experiments, including SpecAugment tuning, clean, and noise-aware fine-tuning, required over 240 GPU-hours.

We also the Large variant to be slightly better than Large-turbo. Whisper large-turbo is a fine-tuned of pruned whisper large. Thus, they are both the exact same model except the turbo variant have reduced number of decoding layers, from 32 to 4. The turbo model is optimized for faster inference with a minor degradation. Therefore, the result we have in Table~\ref{tab:wer-sundanese} and Table~\ref{tab:wer-javanese} is expected since we fine-tune a larger number of parameters in the large variant.

\begin{table}[ht]
\centering
\small
\begin{tabular}{lccc}
\toprule
\textbf{Model} & \textbf{Clean} & \multicolumn{2}{c}{\textbf{Noisy}} \\
\cmidrule(lr){3-4}
              &                & \textbf{+SNR} & \textbf{--SNR}\\
\midrule
\multicolumn{4}{l}{\cellcolor{blue!7} \textbf{Tiny}} \\
Zero-shot            & 128.56 & 170.65 & 205.89 \\
Clean                & \textbf{60.42} & 77.60 & 133.53 \\
SpecAug + Clean      & 60.99 & 78.41 & 133.59\\
NoiseTrain           & 65.09 & \textbf{76.10} & \textbf{106.51}\\
\midrule
\multicolumn{4}{l}{\cellcolor{blue!7} \textbf{Medium}} \\
Zero-shot            & 92.08 & 105.33 & 152.42 \\
Clean                & \textbf{25.40} & 33.85 & 225.38\\
SpecAug + Clean      & 25.45 & 32.79 & 140.05\\
NoiseTrain           & 26.87 & \textbf{32.41} & \textbf{111.89}\\
\midrule
\multicolumn{4}{l}{\cellcolor{blue!7} \textbf{Large-v3}} \\
Zero-shot            & 74.62 & 82.66 & 148.12 \\
Clean                & \textbf{21.14} & 28.47 & 100.76\\
SpecAug + Clean      & 21.45 & 27.45 & \textbf{88.48} \\
NoiseTrain           & 22.50 & \textbf{27.10} & 114.95\\
\midrule
\multicolumn{4}{l}{\cellcolor{blue!7} \textbf{Large-v3-Turbo}} \\
Zero-shot            & 67.13 & 80.29 & 195.65 \\
Clean                & 24.12 & 77.80 & \textbf{134.19} \\
SpecAug + Clean      & \textbf{23.89} & 31.75 & 140.82 \\
NoiseTrain           & 24.79 & \textbf{30.95} & 153.73  \\
\bottomrule
\end{tabular}
\caption{WER on the Javanese test set across clean and noisy conditions. All models are fine-tuned on Javanese only. ``+SNR'' refers to high SNR and ``--SNR'' to low SNR. Zero-shot results are only evaluated on clean audio.}
\label{tab:wer-javanese}
\end{table}

\begin{table}[ht]
\centering
\small
\begin{tabular}{lccc}
\toprule
\textbf{Model} & \textbf{Clean} & \multicolumn{2}{c}{\textbf{Noisy}} \\
\cmidrule(lr){3-4}
              &                & \textbf{+SNR} & \textbf{--SNR}\\
\midrule
\multicolumn{4}{l}{\cellcolor{blue!7} \textbf{Tiny}} \\
Zero-shot            & 116.79 & 194.18 & 360.48\\
Clean                & 40.37 & 68.50 & 413.56\\
SpecAug + Clean      & \textbf{40.19} & 61.64 & 274.32\\
NoiseTrain           & 43.82 & \textbf{58.89} & \textbf{201.79}\\
\midrule
\multicolumn{4}{l}{\cellcolor{blue!7} \textbf{Medium}} \\
Zero-shot            & 83.20 & 93.06 & 282.98 \\
Clean                & \textbf{4.03} & 8.43 & 199.09 \\
SpecAug + Clean      & 4.09 & \textbf{7.84} & 165.36 \\
NoiseTrain           & 5.46 & 8.59 & \textbf{56.15}\\
\midrule
\multicolumn{4}{l}{\cellcolor{blue!7} \textbf{Large-v3}} \\
Zero-shot            & 78.90 & 83.62 & 171.76 \\
Clean                & \textbf{3.72} & 6.60 & 79.91\\
SpecAug + Clean      & 3.98 & 6.24 & 67.59\\
NoiseTrain           & 4.10 & \textbf{5.88} & \textbf{41.37}\\
\midrule
\multicolumn{4}{l}{\cellcolor{blue!7} \textbf{Large-v3-Turbo}} \\
Zero-shot            & 73.20 & 81.04 & 187.01 \\
Clean                & \textbf{4.83} & 9.84 & 160.43\\
SpecAug + Clean      & 4.83 & 8.95 & 124.15\\
NoiseTrain           & 6.17 & \textbf{8.62} & \textbf{65.42}\\
\bottomrule
\end{tabular}
\caption{WER on the Sundanese test set across clean and noisy conditions. All models are fine-tuned on Sundanese only. ``+SNR'' refers to high SNR and ``--SNR'' to low SNR. Zero-shot results are only evaluated on clean audio.}
\label{tab:wer-sundanese}
\end{table}

\subsection{Error Analysis}

We conduct error analysis on the best model, \texttt{Large-v3}, using two views. \textit{First}, we use character error rate (CER) to quantify fine grained edits: extra spaces, vowel changes, consonant changes, and diacritics, which is appropriate for agglutinative languages where small affix or spacing differences can inflate word errors. \textit{Second}, we use WER to summarize word insertions, deletions, and substitutions. Table~\ref{tab:error-distribution-large} reports the CER-based error distribution for Javanese and Sundanese(see Appendix~\ref{appendix:error-analysis}).

\begin{table}[ht]
\centering
\small
\begin{tabular}{lcccc}
\toprule
\multirow{2}{*}{\textbf{Error Type}} & \multicolumn{2}{c}{\textbf{Cased}} & \multicolumn{2}{c}{\textbf{Uncased}} \\
\cmidrule(lr){2-3} \cmidrule(lr){4-5}
 & \textbf{jav} & \textbf{sun} & \textbf{jav} & \textbf{sun} \\
\midrule
Additional Space    & 900  & 338 & 918  & 351 \\
Consonant Mistake       & 7702 & 2284 & 5815 & 1952 \\
Vowel Mistake   & 3722 & 1214  & 3660 & 1236  \\
Diacritics Mistake  & 1702 & 4  & 1680 & 4 \\
\bottomrule
\end{tabular}
\caption{Distribution of different types of errors for Javanese (jav) and Sundanese (sun) language datasets.}
\label{tab:error-distribution-large}
\end{table}

\paragraph{Additional Space}
This error occurs when the model inserts or removes spaces incorrectly. In Javanese, examples include \textit{dipunpanggihaken} becoming \textit{dipun panggihaken}, or \textit{adipati} split into \textit{adi pati}. In Sundanese, errors often involve foreign names (e.g., \textit{baekhyun} → \textit{baek hyun}) or place names (e.g., \textit{situ lengkong} → \textit{situlengkong}). Common words like \textit{minangka} were also occasionally split into \textit{minang ka}.

\paragraph{Vowel Mistakes}
Vowel-related errors often arise from subtle phonetic variations and orthographic influences. In Sundanese, confusion among the three \textit{e}-like vowels—\textipa{e} (as in lebak), \textipa{è} (bèbèk), and \textipa{eu} (teuas)—frequently leads to transcription mistakes, such as heulang being rendered as helang. Foreign names are also problematic when pronounced with local phonology, e.g., Taylor pronounced as Tayler /\textipa{["taj.ler]}/. In Javanese, vowel shifts and reductions are common, with examples like permata becoming permato or terus shortened to trus, reflecting dialectal or colloquial speech that ASR models struggle to handle. Additionally, Dutch-influenced spellings, such as oe for /u/—, can cause errors like Doel being transcribed as Dul.

\paragraph{Consonant Mistake}
These were far more common in Javanese, probably because it has more complex consonant sounds, including digraphs like \textit{dh}, \textit{ng}, \textit{ny}, and \textit{th}, which are sometimes simplified or misheard. Some Javanese examples include \textit{cetha} becoming \textit{ceto}, \textit{baut} as \textit{baud}, \textit{djoni} as \textit{jani}, \textit{aktris} as \textit{apris}, and \textit{putuku} written as \textit{puduku}. In Sundanese, consonant errors were less frequent, but often appeared in borrowed or foreign words. For instance, some speakers pronounce \textit{f} or \textit{v} as \textit{p}, resulting in words like \textit{felton} → \textit{pelton}, \textit{pevita} → \textit{fevita}, or \textit{shidqia} → \textit{shidgya}.

\paragraph{Diacritics Mistake}
Diacritic-related errors were mainly happen in Javanese. Javanese uses diacritics more extensively, especially marks like \textit{é} and \textit{è}, which affect pronunciation and meaning. These are known as \textit{sandhangan swara}. We found examples like \textit{dhèwèké} written as \textit{dhaweke}, \textit{radén} as \textit{radenma}, \textit{warnané} as \textit{warnane}, and \textit{saliyané} as \textit{saliyane}. Additionally, we would like to note that data from OpenSLR in Sundanese does not include diacritics, even though diacritics are supposed to be used in Sundanese to differentiate e and \textit{è} (pronounced differently). Due to the absence of diacritics in the Sundanese transcript, we only observed a few minor cases, involving only the name \textit{Beyoncé}, which was predicted without the accent as \textit{Beyonce}, since the models are fine-tuned without any diacritics.

\section{Limitations}
This study has three main limitations. First, the OpenSLR corpora were only from limited regions, which may not reflect spontaneous or dialectal variation in Javanese and Sundanese. Second, the noisy conditions are synthetic and cannot fully capture real-world environments such as conversational overlap or varied recording devices. Third, our experiments focus only on Whisper-based models with a small set of fine-tuning strategies. These factors constrain the generalizability of the findings but also motivate directions for improvement.

\section{Conclusion}
We evaluated Whisper-based ASR models on Javanese and Sundanese under noisy conditions. While clean audio performance was strong, WER degraded by 2–3× in low-SNR scenarios without noise-aware training. Both \texttt{SpecAugment} and synthetic noise improved robustness, with \texttt{NoiseTrain} consistently outperforming other methods on average across models and languages. Error analysis showed Sundanese struggled with vowel confusion and name errors, while Javanese had more digraph and consonant issues, resulting in higher WER. Future work includes dialect-aware fine-tuning and speech enhancement for better real-world robustness.

\newpage
\bibliography{anthology,custom}
\bibliographystyle{acl_natbib}

\appendix
\onecolumn

\section{Experimental Configuration}
\label{appendix:experiments}
To find the best SpecAugment setup for our training, we ran a series of controlled experiments using different time and frequency masking combinations. Table~\ref{tab:specaug-configs} lists the configurations we tested, each with different masking probabilities, lengths, and minimum number of masks applied to the time and frequency dimensions of the input spectrograms.

We started with individual masking strategies and then explored balanced and mixed configurations. These ranged from light to aggressive settings to see how much augmentation the model could benefit from before performance started to drop. Based on the validation WER, the best-performing configuration was then used to retrain the final model on the whole training set.

\begin{table*}[ht]
\centering
\small
\begin{tabular}{clcccccc}
\toprule
\textbf{Exp} & \textbf{Description} & \textbf{Time Prob} & \textbf{Time Len} & \textbf{Time Min} & \textbf{Freq Prob} & \textbf{Freq Len} & \textbf{Freq Min} \\
\midrule
0 & Baseline (no SpecAugment)      & 0.00 & 0  & 0 & 0.00 & 0  & 0 \\
1 & Light Time Masking Only        & 0.05 & 10 & 2 & 0.00 & 0  & 0 \\
2 & Medium Time Masking Only       & 0.10 & 15 & 2 & 0.00 & 0  & 0 \\
3 & Heavy Time Masking Only        & 0.20 & 20 & 3 & 0.00 & 0  & 0 \\
4 & Light Frequency Masking Only   & 0.00 & 0  & 0 & 0.05 & 10 & 1 \\
5 & Medium Frequency Masking Only  & 0.00 & 0  & 0 & 0.10 & 15 & 2 \\
6 & Balanced Light (Time + Freq)   & 0.05 & 10 & 2 & 0.05 & 10 & 1 \\
7 & Balanced Medium (Time + Freq)  & 0.10 & 12 & 2 & 0.10 & 12 & 2 \\
8 & Time-Heavy Mix                 & 0.15 & 15 & 3 & 0.05 & 8  & 1 \\
9 & Frequency-Heavy Mix            & 0.05 & 8  & 1 & 0.15 & 15 & 3 \\
10 & Aggressive (Heavy Time + Freq) & 0.20 & 20 & 3 & 0.15 & 18 & 3 \\
\bottomrule
\end{tabular}
\caption{SpecAugment configurations used in each experiment. Values represent the masking probabilities, lengths, and minimum number of time and frequency dimensions masks.}
\label{tab:specaug-configs}
\end{table*}

\section{Synthetic Noise Generation}
\label{appendix:noise-generation}

To simulate real-world conditions, we create a set of noisy training data by mixing clean speech from the OpenSLR dataset with different types of background noise. We follow the general approach of \citet{orel2023noise} and use samples from \texttt{AudioSet} as our noise source. The noise types we picked were meant to reflect various environments in which people often speak, such as traffic, crowds, or indoor chatter, listed in Appendix~\ref{appendix:audioset}.

In our experiments, we use the following Signal-to-Noise Ratio (SNR) values: \texttt{-20, -15, -10, -5, 0, 5, 10, 15, 20, clean}, where \texttt{clean} refers to the original audio without any added noise. Negative SNR values mean more noise relative to the speech, whereas positive values are closer to clean conditions. We specifically chose these values, similar to prior work \cite{maas2012recurrent}, since they cover the full spectrum of acoustic conditions from severe noise corruption to optimal listening environments.

To generate the noisy samples, we use the following formula:\[\texttt{noisy\_audio} = \texttt{original\_audio} + \alpha \cdot \texttt{noise}\]

The scaling factor $\alpha$ controls how much noise is added and is calculated based on the target SNR using:

\[
{\alpha = \sqrt{10^{-\frac{\texttt{SNR}}{10}} \cdot \frac{ \lVert\texttt{original\_audio}\rVert_2^2 }{ \lVert\texttt{noise}\rVert_2^2 }}}
\]

\section{Error Analysis}
\label{appendix:error-analysis}

We analyzed the outputs of all Whisper models to understand the kinds of errors made in Javanese and Sundanese. To focus on more meaningful mistakes, we ignored casing differences.

\subsection{Character-level error analysis (CER)}
We analyze CER to capture small edits common in agglutinative morphology, grouping aligned character edits into four types: extra spaces, vowel errors, consonant errors, and diacritic errors. Table~\ref{tab:cer-error-distribution} reports counts by model and language: Javanese is dominated by consonant and diacritic changes, whereas Sundanese shows relatively more vowel and consonant changes; lowercasing the text (uncased CER) consistently reduces total character edits by about 7–18\% across models, indicating that many mismatches are orthographic rather than full lexical substitutions. For computation, we normalize reference and hypothesis to NFC, collapse repeated whitespace, apply casefolding for uncased scoring, and compute $\mathrm{CER}=\frac{S+D+I}{N}$, where $S$, $D$, and $I$ are minimal character substitutions, deletions, and insertions from the alignment and $N$ is the number of reference characters; error types are assigned from aligned edits: whitespace $\rightarrow$ space; $\{a,i,u,e,o\}\rightarrow$ vowel; base–diacritic pairs (e.g., $e$ vs.\ \'{e}) $\rightarrow$ diacritics; remaining letters $\rightarrow$ consonant.

\begin{table}[ht]
\centering
\small
\setlength{\tabcolsep}{5pt}
\begin{tabular}{lcccccccc}
\toprule
\textbf{Error Type} & \multicolumn{2}{c}{\textbf{Tiny}} & \multicolumn{2}{c}{\textbf{Medium}} & \multicolumn{2}{c}{\textbf{Large-v3}} & \multicolumn{2}{c}{\textbf{Large-v3-turbo}} \\
\cmidrule(lr){2-3} \cmidrule(lr){4-5} \cmidrule(lr){6-7} \cmidrule(lr){8-9}
 & \textbf{jav} & \textbf{sun} & \textbf{jav} & \textbf{sun} & \textbf{jav} & \textbf{sun} & \textbf{jav} & \textbf{sun} \\
\midrule
\rowcolor{blue!10}
\multicolumn{9}{c}{\textbf{Cased}} \\
Additional space        & 6249 & 6110 & 1278 & 419 & 900 & 338 & 1039 & 391 \\
Consonant mistake   & 32881 & 30614 & 9611 & 2552 & 7702 & 2284 & 8632 & 2810 \\
Vowel mistake       & 15744 & 14417 & 4742 & 1494 & 3722 & 1214 & 4168 & 1563 \\
Diacritics mistake  & 3343 & 8 & 1797 & 0 & 1702 & 4 & 1799 & 1 \\
\midrule
\rowcolor{blue!10}
\multicolumn{9}{c}{\textbf{Uncased}} \\
Additional Space        & 6355 & 6402 & 1308 & 439 & 918 & 351 & 1057 & 391 \\
Consonant mistake    & 26693 & 21061 & 7157 & 2135 & 5815 & 1952 & 6392 & 2408 \\
Vowel mistake        & 15650 & 14606 & 4661 & 1416 & 3660 & 1236 & 4119 & 1578 \\
Diacritics mistake  & 3300 & 7 & 1759 & 0 & 1680 & 4 & 1793 & 1 \\
\midrule
\textbf{Reduction (\%)} & 10.68 & 17.65 & 14.56 & 8.19 & 13.88 & 7.45 & 14.52 & 7.48 \\
\bottomrule
\end{tabular}
\caption{Character-level error type counts for Javanese (jav) and Sundanese (sun) across model sizes under cased and uncased evaluation; the bottom row shows the relative CER reduction (\%) from cased to uncased per column.}
\label{tab:cer-error-distribution}
\end{table}

\subsection{Word-level error analysis (WER)}
We decompose word errors into insertions (I), deletions (D), and substitutions (S) under cased and uncased scoring, Table~\ref{tab:wer-error-breakdown} reports per-language counts across model sizes, and the bottom row gives the relative reduction in total word edits when lowercasing is applied. For computation, we normalize reference and hypothesis to NFC, collapse repeated whitespace, apply casefolding for uncased scoring, tokenize by whitespace, and obtain minimal word-level alignments to count \(I\), \(D\), and \(S\); word error rate is then \( \mathrm{WER} = \frac{S + D + I}{N_{\text{ref words}}} \).

\begin{table}[ht]
\centering
\small
\setlength{\tabcolsep}{5pt}
\begin{tabular}{lcccccccc}
\toprule
\textbf{Error Type} & \multicolumn{2}{c}{\textbf{Tiny}} & \multicolumn{2}{c}{\textbf{Medium}} & \multicolumn{2}{c}{\textbf{Large-v3}} & \multicolumn{2}{c}{\textbf{Large-v3-turbo}} \\
\cmidrule(lr){2-3} \cmidrule(lr){4-5} \cmidrule(lr){6-7} \cmidrule(lr){8-9}
 & \textbf{jav} & \textbf{sun} & \textbf{jav} & \textbf{sun} & \textbf{jav} & \textbf{sun} & \textbf{jav} & \textbf{sun} \\
\midrule
\rowcolor{blue!10}
\multicolumn{9}{c}{\textbf{Cased}} \\
Insertion     & 1541 & 1551 & 472 & 105 & 344 & 63  & 376 & 104 \\
Deletion      & 2587 & 2615 & 592 & 224 & 414 & 227 & 526 & 191 \\
Substitution  & 22178 & 17563 & 9995 & 1842 & 8445 & 1713 & 9600 & 2305 \\
\midrule
\rowcolor{blue!10}
\multicolumn{9}{c}{\textbf{Uncased}} \\
Insertion     & 1546 & 1562 & 472 & 105 & 345 & 63  & 377 & 105 \\
Deletion      & 2592 & 2625 & 592 & 224 & 415 & 227 & 527 & 192 \\
Substitution  & 20535 & 15339 & 8715 & 1767 & 7386 & 1640 & 8359 & 2208 \\
\midrule
\textbf{Reduction (\%)} & 6.21 & 10.13 & 11.57 & 3.45 & 11.49 & 3.64 & 11.80 & 3.65 \\
\bottomrule
\end{tabular}
\caption{Word-level error type counts (WER components) for Javanese (jav) and Sundanese (sun) across model sizes under cased and uncased evaluation. The bottom row shows the relative reduction (\%) in total word edits per column.}
\label{tab:wer-error-breakdown}
\end{table}

\section{Experimental Result}
\label{appendix:snr-results}

\begin{figure*}[htbp]
    \centering

    \begin{subfigure}[c]{0.3\textwidth}
        \centering
        \includegraphics[width=\linewidth]{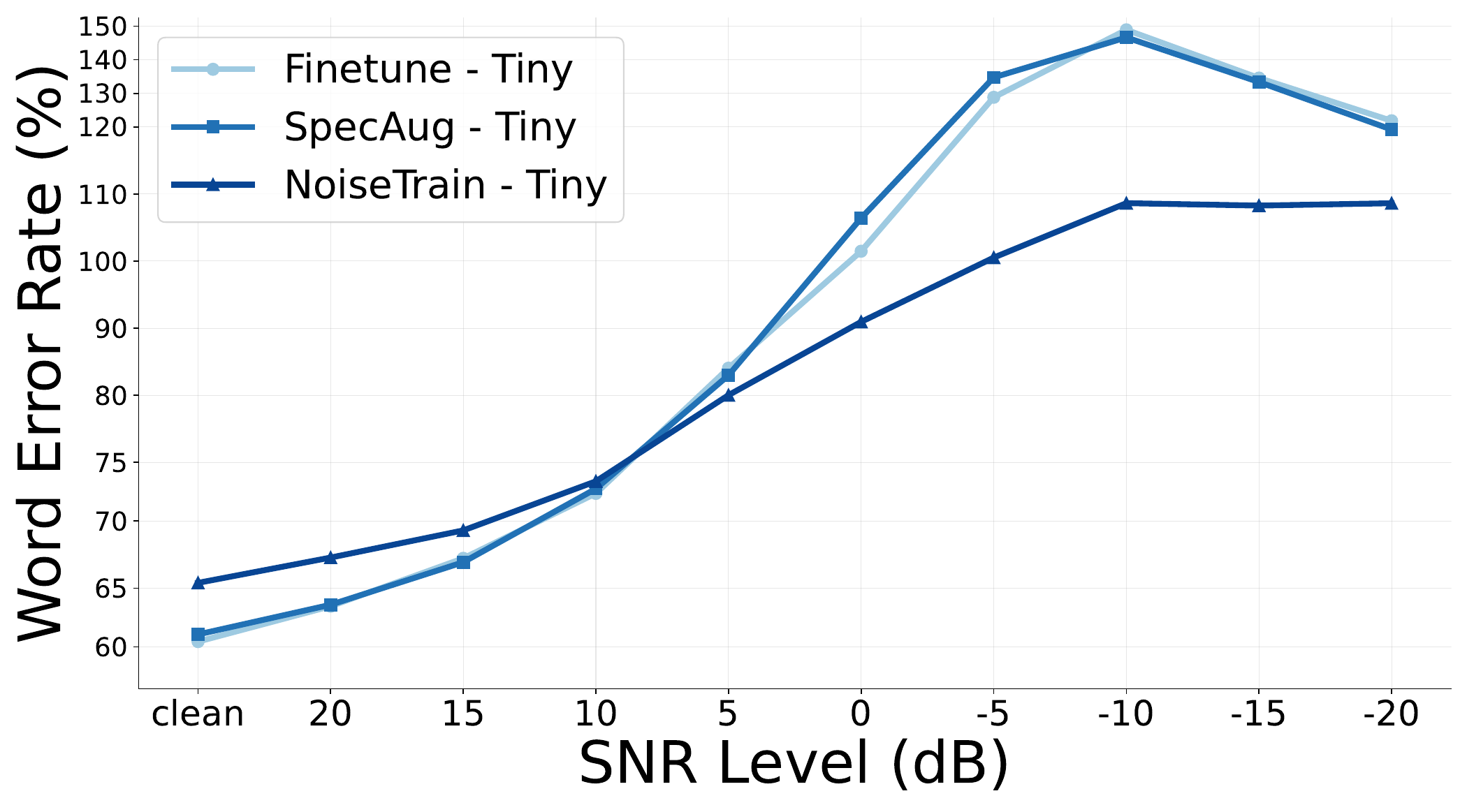}
        \caption{}
        \label{fig:tiny-jv}
    \end{subfigure}
    \hfill
    \begin{subfigure}[c]{0.3\textwidth}
        \centering
        \includegraphics[width=\linewidth]{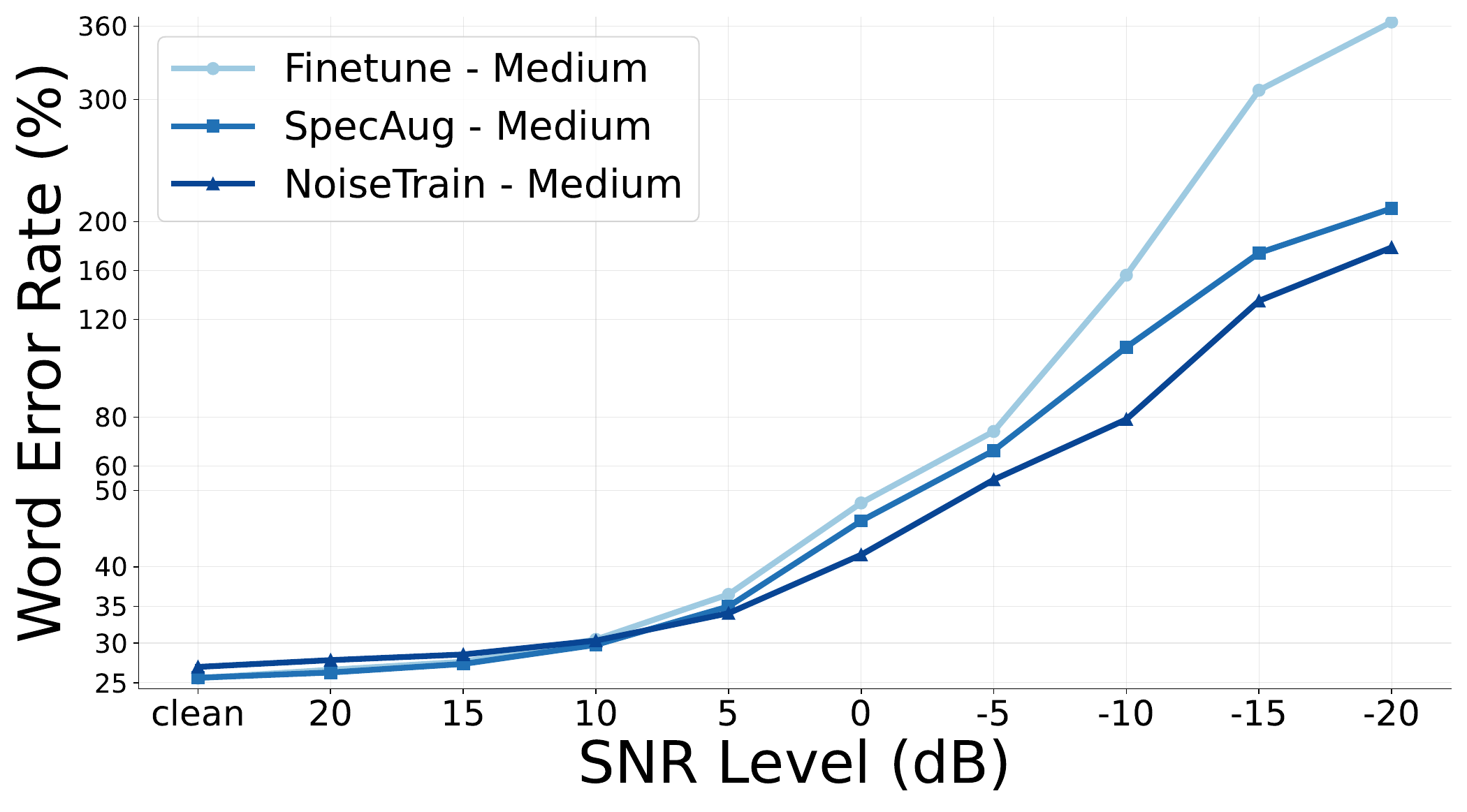}
        \caption{}
        \label{fig:med-jv}
    \end{subfigure}
    \hfill
    \begin{subfigure}[c]{0.3\textwidth}
        \centering
        \includegraphics[width=\linewidth]{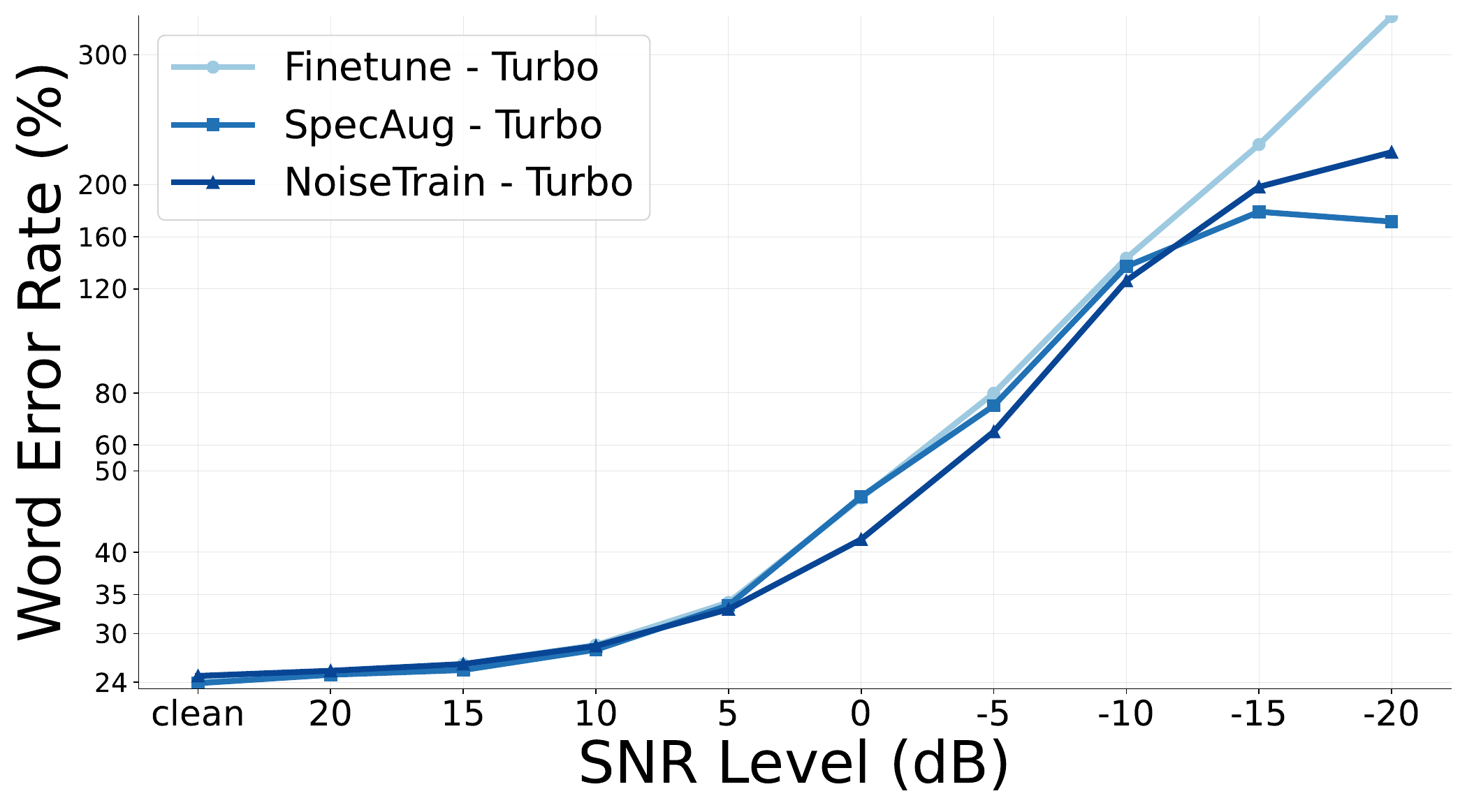}
        \caption{}
        \label{fig:large-jv}
    \end{subfigure}

    \vspace{1em}

    \begin{subfigure}[c]{0.3\textwidth}
        \centering
        \includegraphics[width=\linewidth]{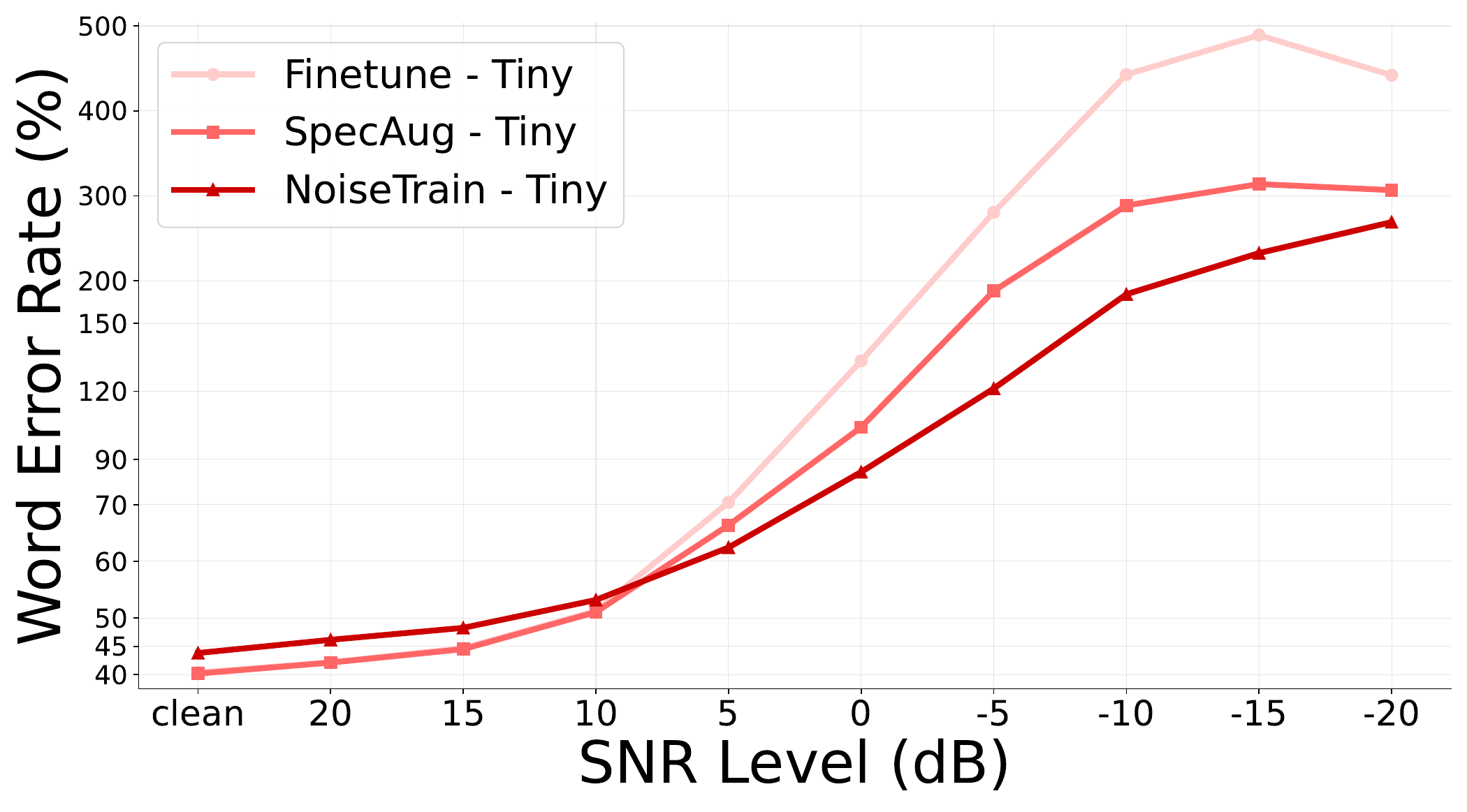}
        \caption{}
        \label{fig:tiny-su}
    \end{subfigure}
    \hfill
    \begin{subfigure}[c]{0.3\textwidth}
        \centering
        \includegraphics[width=\linewidth]{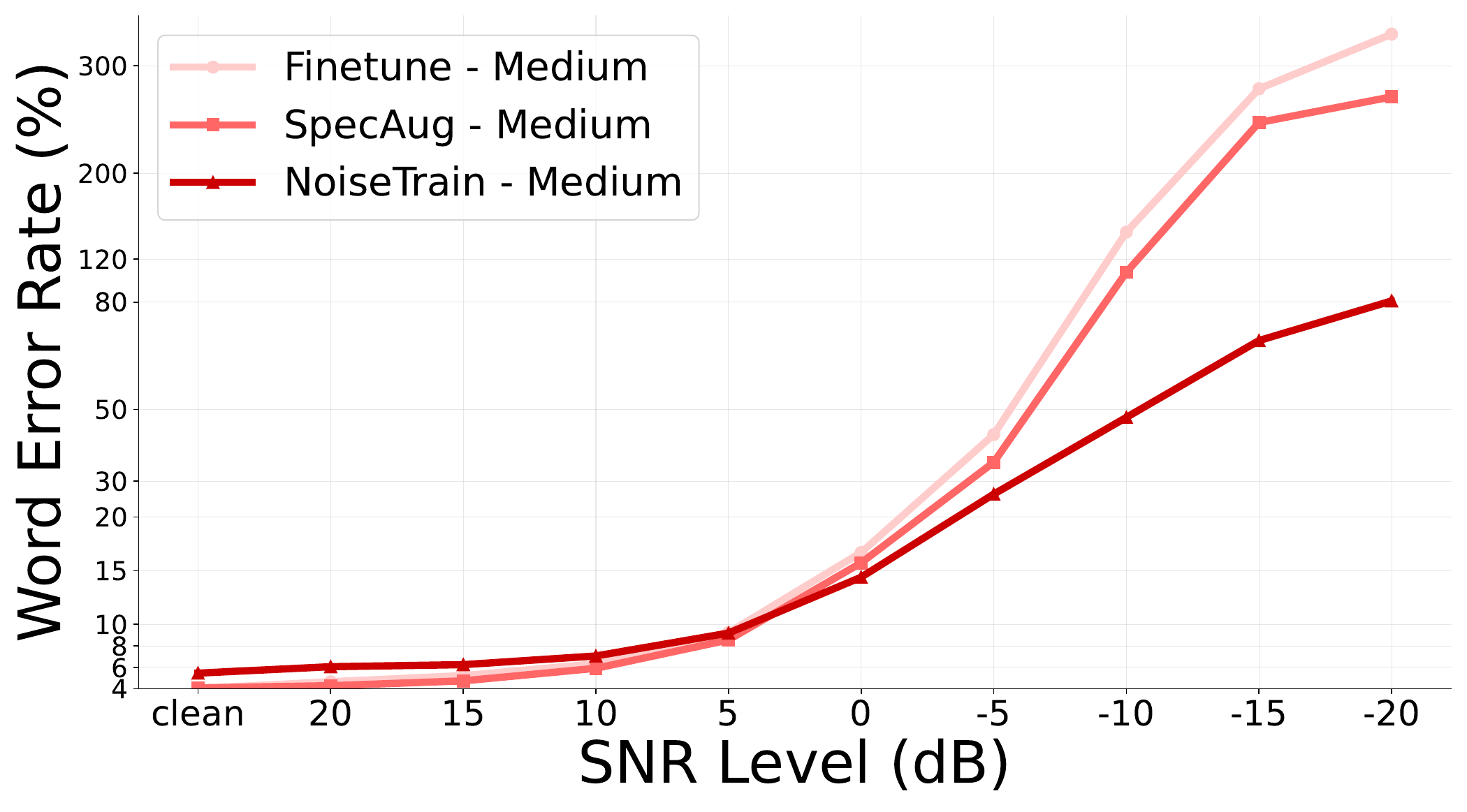}
        \caption{}
        \label{fig:med-su}
    \end{subfigure}
    \hfill
    \begin{subfigure}[c]{0.3\textwidth}
        \centering
        \includegraphics[width=\linewidth]{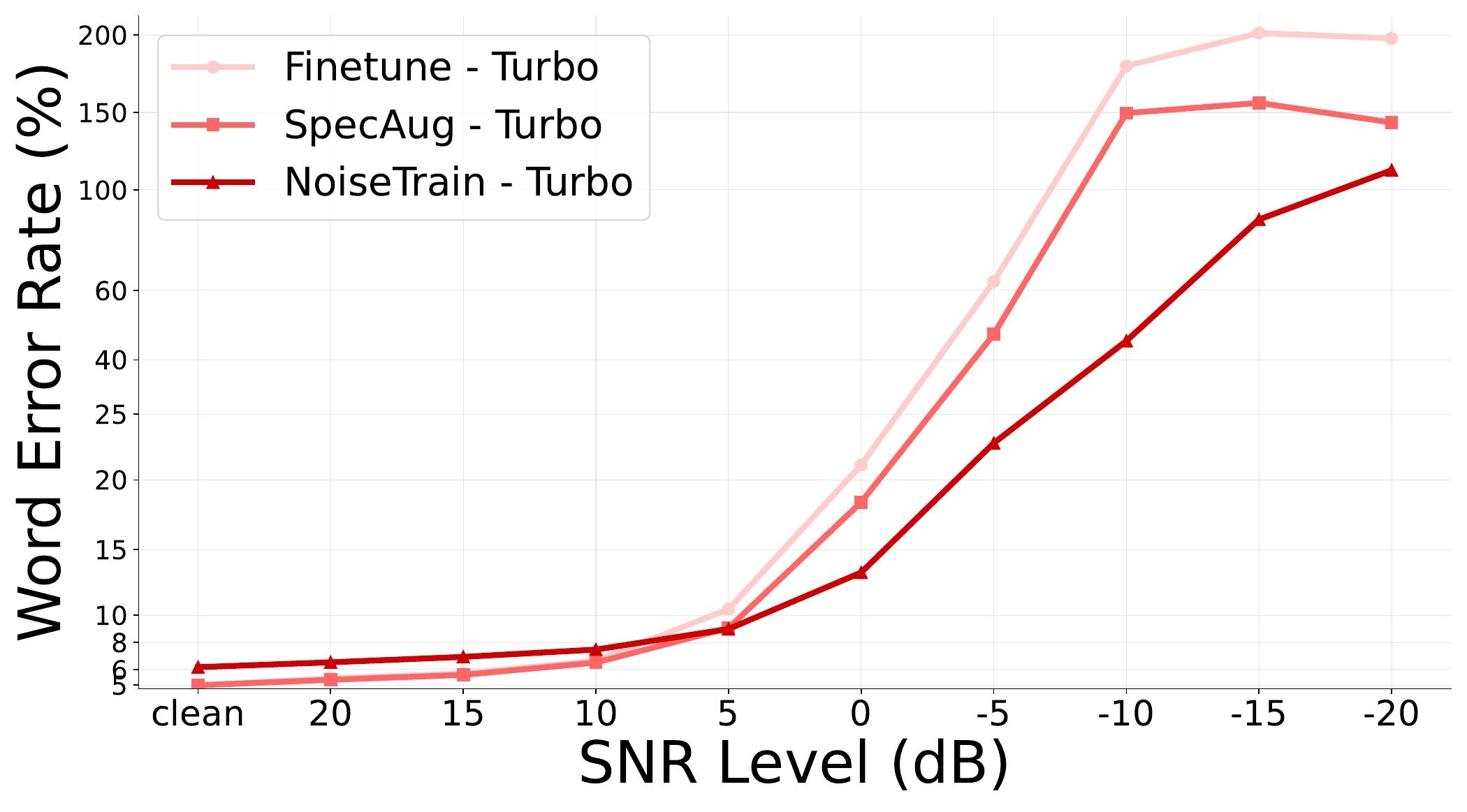}
        \caption{}
        \label{fig:large-su}
    \end{subfigure}

    \caption{WER performance of Whisper variants across different SNR levels for Javanese and Sundanese: (a) Tiny - Javanese, (b) Medium - Javanese, (c) Large-v3-Turbo - Javanese, (d) Tiny - Sundanese, (e) Medium - Sundanese, (f) Large-v3-Turbo - Sundanese.}
    \label{fig:whisper-wer-grid2}
\end{figure*}

We report WER across SNR levels in Tables~\ref{tab:snr-javanese} and~\ref{tab:snr-sundanese} and visualize the trends in Fig.~\ref{fig:whisper-wer-grid2}. The tables cover four Whisper variants (\texttt{Tiny}, \texttt{Medium}, \texttt{Large-v3}, \texttt{Large-v3-Turbo}), each trained with \textbf{Clean}, \textbf{SpecAug+Clean}, and \textbf{NoiseTrain}. Figure~\ref{fig:whisper-wer-grid2} shows \texttt{Tiny}, \texttt{Medium}, and \texttt{Large-v3-Turbo} for both languages, and Figure~\ref{fig:whisper-wer-grid} presents the \texttt{Large-v3} curves. \textbf{As expected, WER increases as SNR decreases, and smaller models degrade more. Noise aware training reduces this drop, especially at low SNR.}

\begin{table*}[htbp!]
\centering
\small
\begin{tabular}{lcccccccccc}
\toprule
\multicolumn{11}{c}{\cellcolor{blue!7} \textbf{Tiny}} \\
\midrule
\textbf{Model} & -20 & -15 & -10 & -5 & 0 & 5 & 10 & 15 & 20 & Clean \\
\midrule
Clean                & 121.82 & 134.56 & 148.93 & 128.82 & 101.47 & 84.04 & 72.20 & 67.04 & 63.23 & 60.43 \\
SpecAug + Clean      & 119.60 & 133.37 & 146.66 & 134.74 & 106.40 & 82.98 & 72.57 & 66.73 & 63.35 & 60.99 \\
NoiseTrain           & 108.62 & 108.27 & 108.63 & 100.52 & 90.94 & 80.02 & 73.16 & 69.26 & 67.10 & 65.09 \\
\midrule
\multicolumn{11}{c}{\cellcolor{blue!7} \textbf{Medium}} \\
\midrule
\textbf{Model} & -20 & -15 & -10 & -5 & 0 & 5 & 10 & 15 & 20 & Clean \\
\midrule
Clean                & 363.54 & 307.64 & 156.21 & 74.14 & 48.35 & 36.37 & 30.47 & 27.55 & 26.50 & 25.40 \\
SpecAug + Clean      & 210.88 & 174.37 & 108.62 & 66.32 & 46.00 & 34.79 & 29.76 & 27.26 & 26.14 & 25.45 \\
NoiseTrain           & 178.90 & 135.18 & 79.08  & 54.38 & 41.56 & 33.90 & 30.32 & 28.50 & 27.76 & 26.87 \\
\midrule
\multicolumn{11}{c}{\cellcolor{blue!7} \textbf{Large-v3}} \\
\midrule
\textbf{Model} & -20 & -15 & -10 & -5 & 0 & 5 & 10 & 15 & 20 & Clean \\
\midrule
Clean                & 119.07 & 122.72 & 98.30  & 62.93 & 41.02 & 30.40 & 25.48 & 23.27 & 22.16 & 21.14 \\
SpecAug + Clean      &  108.07      & 102.49 & 86.18  & 57.19 & 38.08 & 29.06 & 24.93 & 22.95 & 22.21 & 21.45 \\
NoiseTrain    & 219.07 &123.84&68.47&48.41&35.18&28.37&25.16 &23.73 &23.06&22.50\\
\midrule
\multicolumn{11}{c}{\cellcolor{blue!7} \textbf{Large-v3-Turbo}} \\
\midrule
\textbf{Model} & -20 & -15 & -10 & -5 & 0 & 5 & 10 & 15 & 20 & Clean \\
\midrule
Clean                & 146.75 & 147.43 & 137.47 & 105.12 & 89.64 & 79.23 & 75.50 & 72.72 & 71.89 & 24.12 \\
SpecAug + Clean      &171.69&179.26&137.23&75.10&46.84&33.47&28.01&25.51& 24.93       &     23.89   \\
NoiseTrain           & 225.05 & 198.41 & 126.32 & 65.12  & 41.59 & 32.99 & 28.48 & 26.25 & 25.42 & 24.79 \\
\bottomrule
\end{tabular}
\caption{WER across SNR levels for Javanese}
\label{tab:snr-javanese}
\end{table*}

\begin{table*}[htbp!]
\centering
\small
\begin{tabular}{lcccccccccc}
\toprule
\multicolumn{11}{c}{\cellcolor{blue!7} \textbf{Tiny}} \\
\midrule
\textbf{Model} & -20 & -15 & -10 & -5 & 0 & 5 & 10 & 15 & 20 & Clean \\
\midrule
Clean                & 441.91 & 489.04 & 442.65 & 280.62 & 133.43 & 70.98 & 51.23 & 44.68 & 42.19 & 40.37 \\
SpecAug + Clean      & 306.70 & 313.82 & 288.70 & 188.06 & 104.14 & 66.37 & 51.04 & 44.50 & 42.13 & 40.19 \\
NoiseTrain           & 269.09 & 232.61 & 184.20 & 121.24 & 84.40 & 62.44 & 53.18 & 48.26 & 46.15 & 43.82 \\
\midrule
\multicolumn{11}{c}{\cellcolor{blue!7} \textbf{Medium}} \\
\midrule
\textbf{Model} & -20 & -15 & -10 & -5 & 0 & 5 & 10 & 15 & 20 & Clean \\
\midrule
Clean                & 329.53 & 278.64 & 145.19 & 43.01 & 16.69 & 9.23 & 6.33 & 5.23 & 4.66 & 4.03 \\
SpecAug + Clean      & 271.12 & 247.21 & 107.82 & 35.27 & 15.73 & 8.55 & 5.91 & 4.74 & 4.29 & 4.09 \\
NoiseTrain           & 81.13  & 69.27  & 47.81  & 26.37 & 14.39 & 9.19 & 7.06 & 6.25 & 6.06 & 5.46 \\
\midrule
\multicolumn{11}{c}{\cellcolor{blue!7} \textbf{Large-v3}} \\
\midrule
\textbf{Model} & -20 & -15 & -10 & -5 & 0 & 5 & 10 & 15 & 20 & Clean \\
\midrule
Clean                & 107.27 & 100.49 & 79.05  & 32.81 & 12.80 & 7.08 & 4.94 & 4.26 & 3.91 & 3.72 \\
SpecAug + Clean      &   96.16    &87.53&61.77&24.88&11.16&6.49&4.99&      &4.40&4.14\\
NoiseTrain           &  65.07  & 51.02  & 32.23  & 17.15 & 9.30  & 6.16 & 5.07 & 4.54 & 4.31 & 4.10 \\
\midrule
\multicolumn{11}{c}{\cellcolor{blue!7} \textbf{Large-v3-Turbo}} \\
\midrule
\textbf{Model} & -20 & -15 & -10 & -5 & 0 & 5 & 10 & 15 & 20 & Clean \\
\midrule
Clean                & 197.83 & 201.34 & 180.01 & 62.52 & 21.10 & 10.45 & 6.64 & 5.70 & 5.33 & 4.83 \\
SpecAug + Clean      &143.48&156.14&149.58&47.39&18.35&9.06&6.51&5.60&5.24&   4.83   \\
NoiseTrain           & 112.65 & 80.81  & 45.48  & 22.72 & 13.16 & 8.99 & 7.47 & 6.93 & 6.53 & 6.17 \\
\bottomrule
\end{tabular}
\caption{WER across SNR levels for Sundanese}
\label{tab:snr-sundanese}
\end{table*}

\section{Noise Classes from AudioSet}
\label{appendix:audioset}

\newcommand{\heldoutmarker}{\textsuperscript{*}}

We provide a list in Table~\ref{tab:class_descriptions} of environmental and synthetic noise classes used during training and evaluation, sourced from AudioSet. These include a variety of real-world and synthetic sound events, some of which were used as held-out classes for testing generalization. Held-out classes are marked with a superscript \heldoutmarker.

\begin{longtable}{@{}p{3cm} >{\RaggedRight}p{7cm} r@{}}
\toprule
\textbf{Class Name} & \textbf{Description} & \textbf{Count} \\
\midrule
\endfirsthead

\multicolumn{3}{c}{{\bfseries \tablename~\nexttablenum{} -- continued from previous page}}\\
\toprule
\textbf{Class Name} & \textbf{Description} & \textbf{Count} \\
\midrule
\endhead

\midrule
\multicolumn{3}{r@{}}{Continued on next page} \\
\endfoot

\bottomrule
\caption{Descriptions and counts of noise classes used from AudioSet. Held-out classes are marked with \heldoutmarker.}
\label{tab:class_descriptions}
\endlastfoot

Siren & The sound of a loud noise-making device used to provide warnings to people nearby. A siren typically consists of a single pitch that changes either smoothly or abruptly on timescales around one second. & 2188 \\
Car passing by & The sound of a motorized vehicle as it passes by a listener close to the vehicle's path. The sound may include engine and tire noise and will typically involve a clear build-up and/or decay of intensity as the vehicle approaches and retreats, as well as possible Doppler shift. & 1010 \\
Clatter & An irregular rattling noise, often produced by rapid movement, consisting of a cluster of transient sounds. & 772 \\
White noise & A random, unstructured sound in which the value at any moment provides no information about the value at any other moment. White noise has equal energy in all frequency bands. & 738 \\
Crackle & An irregular sequence of sharp sounds, as from sudden vaporization of liquids trapped in a burning solid, or from a collection of snapping noises. & 662 \\
Wind noise (microphone) & The noise produced when a strong air current passes over a microphone, causing large amplitude local turbulence, normally recorded as mechanical clipping as the microphone element exceeds its limits of linearity. & 548 \\
Environmental noise\heldoutmarker & The combined sounds of transport, industrial, and recreational activities. & 322 \\
Pink noise\heldoutmarker & Unstructured noise whose energy decreases with frequency such that equal amounts of energy are distributed in logarithmic bands of frequency, typically octaves. & 283 \\
Boom\heldoutmarker & A deep prolonged loud noise. & 283 \\
Firecracker & The sound of a small explosive device primarily designed to produce a large amount of noise, especially in the form of a loud bang. & 279 \\
Microwave oven & Sounds made by a kitchen appliance that heats food by exposing it to microwave radiation, including the noise of the fan, rotation mechanism, and microwave source, as well as the alert sound used to indicate that cooking is complete. & 250 \\
Traffic noise, roadway noise & The combined sounds of many motor vehicles traveling on roads. & 196 \\
Air horn, truck horn & The sound of a pneumatic device mounted on large vehicles designed to create an extremely loud noise for signalling purposes. & 161 \\
Hubbub, speech noise, speech babble & Loud, disordered, unintelligible speech noise from many sources. & 146 \\
Static & A crackling or hissing noise caused by electrical interference. & 101 \\
Inside, public space\heldoutmarker & Sounds that appear to have been recorded in a public space such as store, restaurant, or travel terminus, often characterized by both reverberation and continuous background noise. & 98 \\
Rumble & A loud, low-pitched, dull, continuous noise. & 90 \\
Grunt\heldoutmarker & A short low gruff noise, resembling the sound made by animals such as pigs. Specifically refers to humans. & 73 \\
Stomach rumble\heldoutmarker & A rumbling, growling or gurgling noise produced by movement of the contents of the gastro-intestinal tract. & 64 \\
Noise & A sound that has no perceptible structure and that typically interferes with the perception of more interesting or important sounds. & 58 \\
Knock & A sharp noise of a rigid surface being struck, usually without damage and deliberately, most often with the knuckles of the hand. & 54 \\
Clang\heldoutmarker & A loud, resonant, discordant noise, as of a large and partly hollow metal structure being struck. & 49 \\
Bang & A brief and loud noise. & 38 \\
Squeak\heldoutmarker & A short, high-pitched noise without a sharp attack. & 27 \\
Creak & A high-pitched noise with a perceptible variation in pitch as a result of pressure being shifted or applied on a surface, most commonly on wood. & 16 \\

\end{longtable}

\end{document}